\documentclass[11pt,a4paper]{article}
\usepackage{times,latexsym}
\usepackage{url}
\usepackage[T1]{fontenc}
\usepackage[utf8]{inputenc}
\usepackage{tipa}
\usepackage{enumerate}

\usepackage[acceptedWithA]{tacl2018v2}
\usepackage{microtype}
\usepackage{graphicx}
\usepackage{subfigure}
\usepackage{booktabs} 
\usepackage{mathtools}
\usepackage{tikz}
\usetikzlibrary{bayesnet}
\usetikzlibrary{arrows}

\usepackage{tabularx}
\usepackage{tabulary}
\usepackage{colortbl}
\usepackage{color}
\usepackage{xcolor}

\usepackage{todonotes}
\makeatletter
\newcommand*\iftodonotes{\if@todonotes@disabled\expandafter\@secondoftwo\else\expandafter\@firstoftwo\fi}
\makeatother

\definecolor{edolime}{rgb}{0.9,1,0.3}

\usepackage{amsmath}
\usepackage{amsfonts}
\usepackage{amssymb}
\usepackage{cuted}

\DeclareMathOperator{\Tr}{tr}

\usepackage{algorithm}
\usepackage[noend]{algorithmic}

\usepackage[noabbrev,capitalise]{cleveref}
\newcolumntype{Y}{>{\centering\arraybackslash}X}

\newcommand\rurl[1]{%
  \href{http://#1}{\nolinkurl{#1}}%
}

\title{Parameter Space Factorization for Zero-Shot Learning \\ across Tasks and Languages}

\newcommand{\ucambridge}{\kappa}
\newcommand{\ethz}{\zeta}
\newcommand{\technion}{\tau}

\author{\bf Edoardo M. Ponti$^{\ucambridge}$, Ivan Vuli\'{c}$^{\ucambridge}$, Ryan Cotterell$^{\ucambridge,\ethz}$, \\ {\bf Marinela Parovi\'{c}$^{\ucambridge}$, Roi Reichart$^{\technion}$, Anna Korhonen$^{\ucambridge}$} \\
$^{\ucambridge}$University of Cambridge~\;~$^\ethz$ETH Z\"urich~\;~$^{\technion}${Technion, IIT} \\
$^{\ucambridge}$\texttt {\{ep490,iv250,rdc42,mp939,alk23\}@cam.ac.uk} \\
$^{\technion}$\texttt {roiri@ie.technion.ac.il}
}

\date{}

\begin{document}

\newcommand{\vtheta}{{\boldsymbol \theta}}
\newcommand{\vphi}{{\boldsymbol \phi}}
\newcommand{\vups}{{\boldsymbol \upsilon}}
\newcommand{\vlambda}{{\boldsymbol \lambda}}
\newcommand{\vnu}{{\boldsymbol \nu}}
\newcommand{\vepsilon}{{\boldsymbol \epsilon}}
\newcommand{\calN}{{\cal N}}
\newcommand{\calO}{{\cal O}}
\newcommand{\xx}{\mathbf{x}}
\newcommand{\KL}{\text{KL}}
\newcommand{\vmu}{{\boldsymbol \mu}}
\newcommand{\vsigma}{{\boldsymbol \sigma}}
\newcommand{\vSigma}{{\boldsymbol \Sigma}}
\newcommand{\defn}[1]{{#1}}
\newcommand{\qlambda}{q_{\vlambda}}
\newcommand{\bert}{\textsc{bert}}
\newcommand{\softmax}{\mathrm{softmax}}
\newcommand{\R}{\mathbb{R}}

\DeclarePairedDelimiterX{\infdivx}[2]{(}{)}{%
  #1\;\delimsize\|\;#2%
}
\newcommand{\infdiv}{\mathbb{KL}\infdivx}
\newcommand{\lv}[1]{{\boldsymbol #1}}
\newcommand{\KLD}[2]{\mathrm{KL} \left( #1 \mid\mid #2 \right) }

\maketitle

\begin{abstract}
    Most combinations of NLP tasks and language varieties lack in-domain examples for supervised training because of the paucity of annotated data. How can neural models make sample-efficient generalizations from task--language combinations with available data to low-resource ones? In this work, we propose a Bayesian generative model for the space of neural parameters. We assume that this space can be factorized into latent variables for each language and each task. We infer the posteriors over such latent variables based on data from seen task--language combinations through variational inference. This enables zero-shot classification on unseen combinations at prediction time. For instance, given training data for named entity recognition (NER) in Vietnamese and for part-of-speech (POS) tagging in Wolof, our model can perform accurate predictions for NER in Wolof. 
    In particular, we experiment with a typologically diverse sample of 33 languages from 4 continents and 11 families, and show that our model yields comparable or better results than state-of-the-art, zero-shot cross-lingual transfer methods. 
    Moreover, we demonstrate that approximate Bayesian model averaging results in smoother predictive distributions, whose entropy inversely correlates with accuracy. Hence, the proposed framework also offers robust estimates of prediction uncertainty.
    Our code is located at \rurl{github.com/cambridgeltl/parameter-factorization}.
\end{abstract}

\section{Introduction}
\label{s:intro}
The annotation efforts in NLP have achieved impressive feats, such as the Universal Dependencies (UD) project \cite{UD2.4} which now includes 83 languages. 
But, even UD covers only a meager subset of the world's estimated 8,506 languages \cite{glottolog}. 
What is more, Wikipedia\footnote{\rurl{en.wikipedia.org/wiki/Category:Tasks_of_natural_language_processing}} lists 32 separate NLP tasks.
The lack of costly and labor-intensive labeled data for many of such task--language combinations hinders the development of computational models for the majority of the world's languages \cite{Snyder:2010icml,Ponti:2019cl}.

A common solution is transferring 
knowledge across domains, such as tasks and languages \cite{yogatama2019learning,Talmor:2019acl}, which holds promise to mitigate the lack of training data inherent to a large spectrum of NLP applications \cite[\textit{inter alia}]{Tackstrom:2012naacl,Agic:2016tacl,Ammar:2016tacl,Ponti:2018acl,Ziser:2018emnlp}. 
In the most extreme scenario, \textit{zero-shot learning}, no annotated examples are available for the target domain. In particular, zero-shot transfer across \textit{languages} implies a change in the data domain, and leverages information from resource-rich languages to tackle the same task in a previously unseen target language \cite[\textit{inter alia}]{Lin:2019acl,Rijhwani:2019aaai,Artetxe:2019tacl,Ponti:2019cl}. 
Zero-shot transfer across \textit{tasks} within the same language \cite{ruder2019transfer}, on the other hand, implies a change in the space of labels.

As our main contribution, we propose a Bayesian generative model of the neural parameter space. We assume this to be structured, and for this reason factorizable into task- and language-specific latent variables.\footnote{By latent variable we mean every variable that has to be inferred from observed (directly measurable) variables. To avoid confusion, we use the terms \textit{seen} and \textit{unseen} when referring to different task--language combinations.} By performing transfer of knowledge from both related tasks \textit{and} related languages (i.e., from \textit{seen} combinations), our model allows for zero-shot prediction on \emph{unseen} task--language combinations. 
For instance, the availability of annotated data for part-of-speech (POS) tagging in Wolof and for named-entity recognition (NER) in Vietnamese supplies plenty of information to infer a task-agnostic representation for Wolof and a language-agnostic representation for NER. Conditioning on these, the appropriate neural parameters for Wolof NER can be generated at evaluation time.
While this idea superficially resembles matrix completion for collaborative filtering \cite[]{mnih2008probabilistic,dziugaite2015neural}, the neural parameters are latent and are non-identifiable. Rather than recovering missing entries from partial observations, in our approach we reserve latent variables to each language and each task to tie together neural parameters for combinations that have either of them in common.

We adopt a Bayesian perspective towards inference.
The posterior distribution over the model's latent variables is approximated through stochastic variational inference \cite[SVI;][]{hoffman2013stochastic}.
Given the enormous number of parameters, we also explore a memory-efficient inference scheme based on a diagonal plus low-rank approximation of the co-variance matrix. This guarantees that our model remains both expressive and tractable.

We evaluate the model on two sequence labeling tasks: POS tagging and NER, relying on a typologically representative sample of 33 languages from 4 continents and 11 families.
The results clearly indicate that our generative model surpasses standard baselines based on cross-lingual transfer 1) from the (typologically) nearest source language; 2) from the source language with the most abundant in-domain data (English); and 3) from multiple source languages, in the form of either a multi-task, multi-lingual model with parameter sharing \citep{wu2019beto} or an ensemble of task- and language-specific models \citep{rahimi2019massively}. 

Finally, we empirically demonstrate the importance of modeling uncertainty during inference through Monte Carlo approximations of Bayesian model averaging. This endows neural networks with the ability to ``fail loudly'' \cite{rabanser2019failing} in low-confidence settings such as zero-shot cross-lingual and cross-task transfer. As a result, our generative model enhances both accuracy and robustness in low-resource NLP tasks.

\section{Bayesian Generative Model}
\label{s:methodology}
In this work, we propose a Bayesian generative model for multi-task, multi-lingual NLP. We train a single Bayesian neural network for several tasks and languages jointly. Formally, we consider a set $T = \{t_1, \dots, t_n\}$ of $n$ tasks and a set $L = \{l_1, \dots, l_m\}$ of $m$ languages.  The core modeling assumption we make is that the parameter space of the neural network is \emph{structured}: specifically, we posit that certain parameters correspond to tasks and others correspond to languages.
This structure assumption allows us to generalize to unseen task--language pairs. 
In this regard, the model is reminiscent of matrix factorization as applied to collaborative filtering \citep{mnih2008probabilistic,dziugaite2015neural}.

We now describe our generative model in three steps that match the nesting level of the plates in the diagram in \cref{fig:grf}.
Equivalently, the reader can follow the nesting level of the \textbf{for} loops in \cref{alg:genmodel} for an algorithmic illustration of the generative story.

\begin{enumerate}[(1)]

\item \textbf{Sampling Task and Language Representations:}
To kick off our generative process, we first sample a latent representation for each of the tasks and languages from multivariate Gaussians: $\lv{t}_i \sim \mathcal{N}(\vmu_{t_i}, \Sigma_{t_i})\in \mathbb{R}^h$ and $\lv{l}_j \sim \mathcal{N}(\vmu_{l_j}, \Sigma_{l_j}) \in \mathbb{R}^h$, respectively. 
While we present the model in its most general form,
we take $\vmu_{t_i} = \vmu_{l_j} = \mathbf{0}$
and $\Sigma_{t_i} = \Sigma_{l_j} = {I}$ for the experimental portion of this paper.

\item \textbf{Sampling Task--Language-specific Parameters:}
Afterward, to generate task--language-specific neural parameters,
we sample $\vtheta_{ij}$ from $ \mathcal{N}(f_{\lv{\psi}}(\lv{t}_i, \lv{l}_j), \mathrm{diag}(f_{\lv{\phi}}(\lv{t}_i, \lv{l}_j))) \in \mathbb{R}^d$
where  $f_{\lv{\psi}}(\lv{t}_i$, $\lv{l}_j)$ and $f_{\lv{\phi}}(\lv{t}_i$, $\lv{l}_j)$ are learned deep feed-forward neural networks $f_\lv{\psi}: \mathbb{R}^h \rightarrow \mathbb{R}^d $ and $f_\lv{\phi} : \mathbb{R}^h \rightarrow \mathbb{R}_{\geq 0}^d $ parametrized by $\lv{\psi}$ and $\lv{\phi}$, respectively, similar to \citet{kingma2014auto}. These transform the latent representations into the mean $\vmu_{\theta_{ij}}$ and diagonal of the co-variance matrix $\vsigma_{\theta_{ij}}$ for the parameters $\vtheta_{ij}$ associated with $t_i$ and $l_j$. The feed-forward network $f_\lv{\psi}$ just has a final linear layer as the mean can range over $\mathbb{R}^d$ whereas $f_\lv{\phi}$
has a final softplus (defined in \cref{ssec:elbo}) layer to ensure it ranges only over $\mathbb{R}_{\geq 0}^d$.
Following \citet{stoleematrix}, the networks $f_\lv{\psi}$ and $f_\lv{\phi}$ take as input a linear function of the task and language vectors: $\lv{t} \oplus \lv{l} \oplus (\lv{t} - \lv{l}) \oplus (\lv{t} \odot \lv{l})$,
where $\oplus$ stands for concatenation and $\odot$ for element-wise multiplication. The sampled neural parameters $\vtheta_{ij}$ are partitioned into a weight ${W_{ij}} \in \mathbb{R}^{e \times c}$ and a bias $\lv{b_{ij}} \in \mathbb{R}^{c}$, and reshaped appropriately. Hence, the dimensionality of the Gaussian is chosen to reflect the number of parameters in the affine layer, $d = e \cdot c + c$, where  $e$ is the dimensionality of the input token embeddings (detailed in the next paragraph) 
and $c$ is the maximum number of classes across tasks.\footnote{Different tasks might involve different class numbers, the number of parameters hence oscillates. The extra dimensions not needed for a task can be considered as padded with zeros.}
The number of hidden layers and the hidden size of $f_\lv{\psi}$ and $f_\lv{\phi}$ are hyper-parameters discussed in \cref{ssec:hyper}.
We tie the parameters $\lv{\psi}$ and $\lv{\phi}$ for all layers except for the last to reduce the parameter count. 
We note that the space of parameters for all tasks and languages forms a tensor $\Theta \in \R^{n \times m \times d}$, where $d$ is the number of parameters of the largest model.

\begin{figure}
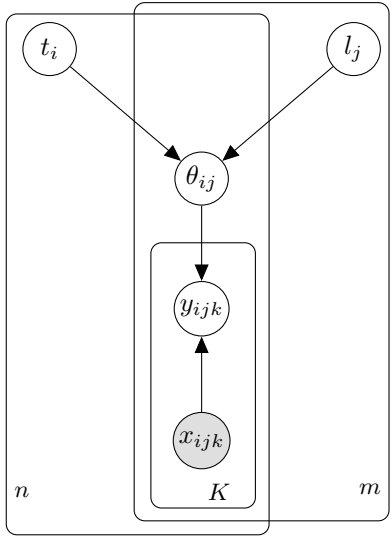

  \centering
  
  \tikz{ %
  \node[latent] (x) {$y_{ijk}$};%
     \node[obs,below=of x] (y) {$x_{ijk}$}; %
     \node[latent,above=of x] (r) {$\theta_{ij}$}; %
     \node[latent,above=of r,xshift=-2cm] (l) {$t_i$}; %
     \node[latent,above=of r,xshift=2cm] (t) {$l_j$}; %
     \plate [inner sep=.3cm,xshift=.02cm,yshift=.2cm] {plate1} {(x) (y)} {$K$}; %
          \plate [inner sep=.3cm,xshift=-0.2cm,yshift=-.05cm] {plate2} {(t) (r) (x) (y)} {$m$}; %
        {
        \tikzset{plate caption/.append style={below=6pt and 0pt of #1.south west}}
         \plate [inner sep=.3cm,xshift=0.2cm,yshift=-.2cm] {plate2} {(l) (r) (x) (y)} {$n$}; %
         }

     \edge {y} {x}
     \edge {r} {x}
     \edge {l,t} {r}
  }

  \caption{A graph (plate notation) of the generative model based on parameter space factorization. Shaded circles refer to observed variables.}
  \label{fig:grf}
\end{figure}

\item \textbf{Sampling Task Labels:}
Finally, we sample the $k^{\text{th}}$ label $y_{ijk}$ for the $i^{\text{th}}$ task and the $j^{\text{th}}$ language from a final softmax: $p(y_{ijk} \mid \xx_{ijk}, \vtheta_{ij}) = \mathrm{softmax}({W}_{ij}\,\bert(\mathbf{x}_{ijk}) + \lv{b}_{ij})$ where $\bert(\mathbf{x}_{ijk}) \in \mathbb{R}^e$ is the multi-lingual BERT \cite{pires2019multilingual} encoder.
The incorporation m-BERT as a pre-trained multlingual embedding allows for enhanced cross-lingual transfer. 

\end{enumerate}

\noindent
Consider the Cartesian product of all tasks and languages $T \times L$.
We can naturally decompose this product into seen task--language pairs $\mathcal{S}$ and unseen task--language pairs $\mathcal{U}$, i.e.
 $T \times L= \mathcal{S} \,\sqcup\, \mathcal{U}$.
Naturally, we are only able to train our model 
on the seen task--language pairs $\mathcal{S}$.
However, as we estimate all task--language parameter vectors $\vtheta_{ij}$ jointly, our model allows us to draw inferences about the parameters for pairs in $\mathcal{U}$ as well. 
The intuition for why this should work is as follows: By observing multiple pairs where the task (language) is the same but the language (task) varies, the model learns to distill the relevant knowledge for zero-shot learning because our generative model structurally enforces a disentangled representations---separating representations for the tasks from the representations for the languages rather than lumping them together into a single entangled representation \cite[\textit{inter alia}]{wu2019beto}.
Furthermore, the neural networks $f_{\lv{\psi}}$ and $f_{\lv{\phi}}$ mapping the task- and language-specific latent variables to neural parameters are shared allowing the model to generalize across task--language pairs.

\renewcommand\algorithmicdo{}
\begin{algorithm}[t]
\caption{Generative Model of Neural Parameters for Multi-task, Multi-lingual NLP.}
   \label{alg:genmodel}
   \begin{algorithmic}[1]
\FOR{$t_i \in T$}{
    \STATE $\lv{t}_i \sim \mathcal{N}(\vmu_{t_i}, \Sigma_{t_i})$ 
}
\ENDFOR
\FOR{$l_j \in L$}{
    \STATE $\lv{l}_j \sim \mathcal{N}(\vmu_{l_j}, \Sigma_{l_j})$ 
}
\ENDFOR
    \FOR{$t_i \in T$}{
        \FOR{$l_j \in L$}{
            \STATE $\vmu_{{\theta}_{ij}} = f_{\lv{\psi}}(\lv{t}_i, \lv{l}_j)$ 
            \STATE $\Sigma_{{\theta}_{ij}} = f_{\lv{\phi}}(\lv{t}_i, \lv{l}_j)$  
            \STATE $\vtheta_{ij} \sim \mathcal{N}(\vmu_{{\theta}_{ij}}, {\Sigma}_{{\theta}_{ij}})$ \\
            
            \FOR{$\xx_{ijk} \in X_{ij}$}{
                \STATE $y_{ijk} \sim p(\cdot \mid \xx_{ijk}, \vtheta_{ij})$
            }
            \ENDFOR
        }
        \ENDFOR
    }
    \ENDFOR
\end{algorithmic}
\end{algorithm}

\def\ci{\perp\!\!\!\perp}

\section{Variational Inference}\label{ssec:elbo}
Exact computation of the posterior over the latent variables $p(\vtheta, \lv{t}, \lv{l} \mid \xx)$ is intractable. 
Thus, we need to resort to an approximation. 
In this work, we consider variational inference as our approximate inference scheme.
Variational inference finds an approximate posterior over the latent variables by minimizing the \defn{variational gap}, which may be expressed as the Kullback--Leibler (KL) divergence between the variational approximation  $q(\vtheta, \lv{t}, \lv{l})$ and the true posterior $p(\vtheta, \lv{t}, \lv{l} \mid \xx)$. 
In our work, we employ the following variational distributions:

\begin{align}
    q_\vlambda &= \calN(\lv{m}_t, { S_t}) \qquad \lv{m}_t \in \mathbb{R}^h, S_t \in \mathbb{R}^{h \times h} \label{eq:qt}\\
    q_\lv{\nu} &= \calN(\lv{m}_l, { S}_l) \qquad \lv{m}_l \in \mathbb{R}^h, S_l \in \mathbb{R}^{h \times h} \label{eq:ql} \\
    q_\lv{\xi} &= \calN({f}_\psi(\lv{t}, \lv{l}), \mathrm{diag}({f}_\phi(\lv{t}, \lv{l}))) \label{eq:tie}
\end{align}
We note the unusual choice to tie parameters between the generative model and the variational family in \cref{eq:tie}; however, we found that this choice performs better in our experiments.

\begin{figure*}[t]
\begin{align}
\KLD{q(\vtheta, \lv{t}, \lv{l})}{p (\vtheta, \lv{t}, \lv{l} \mid \xx)} = - \mathop{\mathbb{E}}_{\lv{t} \sim q_\vlambda} \mathop{\mathbb{E}}_{\lv{l} \sim q_\lv{\nu}} \mathop{\mathbb{E}}_{\vtheta \sim q_\lv{\xi}} \log \frac{p (\vtheta, \lv{t}, \lv{l} \mid \xx)}{q(\vtheta, \lv{t}, \lv{l})} 
= - \mathop{\mathbb{E}}_{\lv{t} \sim q_\vlambda} \mathop{\mathbb{E}}_{\lv{l} \sim q_\lv{\nu}} \mathop{\mathbb{E}}_{\vtheta \sim q_\lv{\xi}} [\log p(\vtheta, \lv{t}, \lv{l}, \xx) \nonumber \\ 
- \log p(\xx) - \log q(\vtheta, \lv{t}, \lv{l})]  = \log p(\xx) - \mathop{\mathbb{E}}_{\lv{t} \sim q_\vlambda} \mathop{\mathbb{E}}_{\lv{l} \sim q_\lv{\nu}} \mathop{\mathbb{E}}_{\vtheta \sim q_\lv{\xi}} \log \frac{p(\vtheta, \lv{t}, \lv{l}, \xx)}{q(\vtheta, \lv{t}, \lv{l})} \triangleq \log p(\xx) - \mathcal{L} \label{eq:gap}
\end{align}
\hrulefill
\end{figure*}

\begin{figure*}[t]
\begin{align}
\log p(\xx) = & \, \log \left( \iiint p(\xx, \vtheta, \lv{t}, \lv{l}) \, \mathrm{d} \vtheta \, \mathrm{d} \lv{t} \, \mathrm{d} \lv{l} \right)  \nonumber \\  
= & \, \log \left( \iiint p(\xx \mid \vtheta) \, p(\vtheta \mid \lv{t}, \lv{l}) \, p(\lv{t}) \, p(\lv{l}) \, \mathrm{d} \vtheta \, \mathrm{d} \lv{t} \, \mathrm{d}\lv{l} \right)  \nonumber \\ 
= & \,\log \left( \iiint \frac{q_\vlambda (\lv{t}) \, q_\lv{\nu}(\lv{l}) \, q_\lv{\xi}(\vtheta \mid \lv{t}, \lv{l})}{q_\vlambda(\lv{t}) \, q_\lv{\nu}(\lv{l}) \, q_\lv{\xi}(\vtheta \mid \lv{t}, \lv{l})} p(\xx \mid \vtheta) \, p(\vtheta \mid \lv{t}, \lv{l}) \, p(\lv{t}) \, p(\lv{l}) \, \mathrm{d} \vtheta \, \mathrm{d} \lv{t} \, \mathrm{d} \lv{l} \right) \nonumber \\ 
= & \, \log \left( \mathop{\mathop{\mathbb{E}}}_{\lv{t} \sim q_\vlambda} \mathop{\mathbb{E}}_{\lv{l} \sim q_\lv{\nu}} \mathop{\mathbb{E}}_{\vtheta \sim q_\lv{\xi}} \frac{ p(\vtheta \mid \lv{t}, \lv{l}) \, p(\lv{t}) \, p(\lv{l})\,p(\xx \mid \vtheta) }{q_\vlambda(\lv{t}) \, q_\lv{\nu}(\lv{l}) \, q_\lv{\xi}(\vtheta \mid \lv{t}, \lv{l})} \right) \nonumber \\ 
\geq  & \, \mathop{\mathbb{E}}_{\lv{t} \sim q_\vlambda} \mathop{\mathbb{E}}_{\lv{l} \sim q_\lv{\nu}} \mathop{\mathbb{E}}_{\vtheta \sim q_\lv{\xi}} \left[ \log \frac{p(\xx \mid \vtheta) \, p(\vtheta \mid \lv{t}, \lv{l}) \, p(\lv{t}) \, p(\lv{l})}{q_\vlambda(\lv{t}) \, q_\lv{\nu}(\lv{l}) \, q_\lv{\xi}(\vtheta \mid \lv{t}, \lv{l})} \right] \triangleq \mathcal{L} \label{eq:elbo} \\ 
= & \, \mathop{\mathbb{E}}_{\lv{t} \sim q_\vlambda} \mathop{\mathbb{E}}_{\lv{l} \sim q_\lv{\nu}} \left[ \mathop{\mathbb{E}}_{\vtheta \sim q_\lv{\xi}} \left[ \log p(\xx \mid \vtheta) + \log \frac{p(\vtheta \mid \lv{t}, \lv{l})}{q_\lv{\xi}(\vtheta \mid \lv{t}, \lv{l})} \right] + \log \frac{p(\lv{t})}{q_\vlambda(\lv{t})} + \log \frac{p(\lv{l})}{q_\lv{\nu}(\lv{l})} \right]  \nonumber \\
= & \,  \underbrace{\mathop{\mathbb{E}}_{\vtheta \sim q_{\lv{\xi}}} \log p(\xx \mid \vtheta)}_\textit{requires approximation} - \underbrace{ \left(\vphantom{\mathop{\mathbb{E}}_{\vtheta \sim q_\lv{\xi}}} \KLD{q_\vlambda(\lv{t})}{p(\lv{t})} + \KLD{q_\lv{\nu}(\lv{l})}{p(\lv{l})} + \KLD{q_\lv{\xi}(\vtheta \mid \lv{t}, \lv{l})}{p(\vtheta \mid \lv{t}, \lv{l})} \right)}_\textit{closed-form solution} \label{eq:2}
\end{align}
\hrulefill
\end{figure*}

Through a standard algebraic manipulation in \cref{eq:gap}, the KL-divergence for our generative model can be shown to equal the marginal log-likelihood $\log p(\xx)$, independent from $q(\cdot)$, and the so-called evidence lower bound (ELBO) $\mathcal{L}$. Thus, approximate inference becomes an optimization problem where maximizing $\mathcal{L}$ results in minimizing
the KL-divergence.
One derives $\mathcal{L}$ is by expanding the marginal log-likelihood as in \cref{eq:elbo} by means of Jensen's inequality. 
We also show that $\mathcal{L}$ can be further broken into a series of terms as illustrated in \cref{eq:2}. 
In particular, we see that it is only the first term in the expansion that requires approximation. The subsequent terms are KL-divergences between variational and true distributions that have closed-form solution due to our choice of prior. 
Due to the parameter-tying scheme above, the KL-divergence in \cref{eq:2} between the variational distribution $q_\lv{\xi}(\vtheta \mid \lv{t}, \lv{l})$ and the prior distribution ${p(\vtheta \mid \lv{t}, \lv{l})}$ is zero. 

In general, the co-variance matrices ${S}_t$ and $S_l$ in \cref{eq:qt} and \cref{eq:ql} will require $\mathcal{O}(h^2)$ space to store. 
As $h$ is often very large, it is impractical to materialize either
matrix in its entirety.
Thus, in this work, we experiment with smaller matrices that have a reduced memory footprint; specifically, we consider a \textit{diagonal} co-variance matrix and a \textit{diagonal plus low-rank} co-variance structure.
A diagonal co-variance matrix makes computation feasible with a complexity of $\calO(h)$; this, however, comes at the cost of not letting parameters influence each other, and thus failing to capture their complex interactions.
To allow for a more expressive variational family, we also consider a co-variance matrix that is the sum of a diagonal matrix and a low-rank matrix:
\begin{align}
    {S}_t &= \mathrm{diag}(\lv{\delta}_t^2) + B_t B_t^\top \\
    {S}_l &= \mathrm{diag}(\lv{\delta}_l^2) + B_l B_l^\top 
\end{align}

\noindent
where ${B} \in \mathbb{R}^{h \times k}$ ensures that $\mathrm{rank}\left(B B^{\top}\right) \leq k$, and $\mathrm{diag}(\lv{\delta})$ is diagonal. 
We can store this structured co-variance matrix in $\mathcal{O}(kh)$ space.

By definition, co-variance matrices must be symmetric and positive semi-definite. The first property holds by construction. 
The second property is enforced by a softplus parameterization where $\mathrm{softplus}(\cdot) \triangleq  \ln (1 + \exp(\lv{\cdot}))$.
Specifically, we define $\lv{\delta}^2 = \mathrm{softplus}(\lv{\rho})$ and we optimize over $\lv{\rho}$. 

\subsection{Stochastic Variational Inference}\label{ssec:svi}
To speed up the training time, we make use of \emph{stochastic} variational inference \cite{hoffman2013stochastic}.
In this setting, we randomly sample a task $t_i \in T$ and language $l_j \in L$ among seen combinations during each training step,\footnote{As an alternative, we experimented with a setup where sampling probabilities are proportional to the number of examples of each task--language combination, but this achieved similar performances on the development sets.} and randomly select a batch of examples from the dataset for the sampled task--language pair.
We then optimize the parameters of the feed-forward neural networks $\lv{\psi}$ and $\lv{\phi}$ as well as the parameters of the variational approximation to the posterior $\lv{m}_t$, $\lv{m}_l$, $\lv{\rho}_t$, $\lv{\rho}_l$, $B_t$ and $B_l$ with a stochastic gradient-based optimizer (discussed in \cref{ssec:hyper}).

The KL divergence terms and their gradients in the ELBO appearing in \cref{eq:2} can be
computed in closed form as the relevant densities are Gaussian \citep[p. 13]{duchi2007derivations}. 
Moreover, they can be calculated for Gaussians with diagonal and low-rank co-variance structures without explicitly unfolding the full matrix. 
For the diagonal co-variance structure:
\begin{align}
\begin{split}
\KLD{q}{p} = &\frac{1}{2} \Bigl[ \sum_{i=1}^{h} (m^2_i + s^2_i) - h - \sum_{i=1}^{h} \ln s_i^2 \Bigr] \label{eq:klnormal}
\end{split}
\end{align}

\noindent
For the low-rank co-variance structure, we have:
\begin{align}
\begin{split}
\KLD{q}{p} = &\frac{1}{2} \Bigl[ \sum_{i=1}^{h} (m^2_i + {\delta}^2_i + \sum_{j=1}^{k} {b}^2_{ij}) \\
&- h - \ln \det ({S}) \Bigr]
\end{split}
\end{align}
where $b_{ij}$ is the element in the $i$-th row and $j$-th column of $B$.
The last term can be estimated without computing the full matrix explicitly thanks to the generalization of the matrix--determinant lemma,\footnote{$ \det(A + UV^\top) = \det(I + V^\top A^{-1} U) \cdot \det(A) $. Note that the lemma assumes that $A$ is invertible.} which, applied to the factored co-variance structure, yields: 
\begin{align}
\begin{split}
    \ln \det({S}) = &\ln \Bigl[ \det({I} + {B}^\top \textrm{diag}({\lv{\delta}^{-2}}) {B}) \Bigr] \\
    & + \sum_{i=1}^h \ln(\delta^2_i)
\end{split}
\end{align}
where ${I} \in \mathbb{R}^k$.

However, as stated before, the following expectation does not admit a closed-form solution. Thus
we consider a Monte Carlo approximation:
\begin{align}
    &\mathop{\mathbb{E}}_{\vtheta \sim q_{\lv{\xi}}}\log p(\xx \mid \vtheta) = \int q_{\lv{\xi}}(\vtheta) \log p(\xx \mid \vtheta)\,\mathrm{d}\vtheta \nonumber \\
    &\approx \sum_{v=1}^V \log p(\xx \mid \vtheta^{(v)}) \quad\text{where} \,\,\,\vtheta^{(v)} \sim q_{\lv{\xi}}  
\end{align}
where $V$ is the number of Monte Carlo samples taken.
In order to allow the gradient to easily flow through the generated samples, we adopt the re-parametrization trick \citep{kingma2014auto}. 
Specifically, we exploit the following identities $ \lv{t}_i = \vmu_{t_i} + \vsigma_{t_i} \odot \lv{\epsilon}$ and $ \lv{l}_j = \vmu_{l_j} + \vsigma_{l_j} \odot \lv{\epsilon}$, where $\lv{\epsilon} \sim \mathcal{N}(\lv{0}, {I})$ and $\odot$ is the Hadamard product. For the diagonal plus low-rank co-variance structure, we exploit the identity:
\begin{equation}
    \vmu + \mathrm{diag}(\lv{\delta}^2 \odot \lv{\epsilon}) + {B} \lv{\zeta}
\end{equation}
\noindent where $\lv{\epsilon} \in \mathbb{R}^{h}$, $\lv{\zeta} \in \mathbb{R}^{k}$, and both are sampled from $\mathcal{N}(\mathbf{0}, {I})$.
The  mean $\vmu_{\theta_{ij}}$ and the diagonal of the co-variance matrix $\vsigma^2_{\theta_{ij}}$ are deterministically
computed given the above samples and the parameters $\vtheta_{ij}$ are sampled from $\mathcal{N}(\vmu_{\theta_{ij}}, \textrm{diag}(\vsigma^2_{\theta_{ij}}))$, 
again with the re-parametrization trick.

\subsection{Computing Posterior Predictive}
During test time, we perform zero-shot predictions on an unseen task--language pair by plugging in the posterior means (under the variational approximation) into the model.
As an alternative, we experimented with ensemble predictions through Bayesian model averaging. 
I.e., for data for seen combinations $\xx_{\mathcal{S}}$ and data for unseen combinations $\xx_{\mathcal{U}}$, the true predictive posterior can be approximated as $p(\xx_\mathcal{U} \mid \xx_\mathcal{S}) = \int p( \xx_\mathcal{U} \mid \vtheta, \xx_\mathcal{S}) \, q_{\lv\xi}(\vtheta \mid \xx_\mathcal{S}) \, \mathrm{d}\vtheta \approx \sum_{v=1}^V p(\xx_\mathcal{U} \mid \vtheta^{(v)}, \xx_\mathcal{S})$, where $V$ are 100 Monte Carlo samples from the posterior $q_\lv{\xi}$. Performances on the development sets are comparable to simply plugging in the posterior mean.

\section{Experimental Setup}
\label{s:experimental}
\subsection{Data}
We select NER and POS tagging as our experimental tasks because their datasets encompass an ample and diverse sample of languages, and are common benchmarks for resource-poor NLP \citep[\textit{inter alia}]{cotterell-duh-2017-low}. In particular, we opt for WikiANN \citep{pan2017cross} for the NER task and Universal Dependencies 2.4 \citep[UD;][]{UD2.4} for POS tagging. Our sample of languages is chosen from the intersection of those available in WikiANN and UD. 
However, we remark that this sample is heavily biased towards the Indo-European family \cite{gerz-etal-2018-relation}. Instead, the selection should be: i) typologically diverse, to ensure that the evaluation scores truly reflect the expected cross-lingual performance \citep{ponti2020xcopa}; ii) a mixture of resource-rich and low-resource languages, to recreate a realistic setting and to allow for studying the effect of data size.
Hence, we further filter the languages in order to make the sample more balanced. In particular, we sub-sample Indo-European languages by including only resource-poor ones, and keep all the languages from other families. 
Our final sample comprises 33 languages from 4 continents (17 from Asia, 11 from Europe, 4 from Africa, and 1 from South America) and from 11 families (6 Uralic, 6 Indo-European, 5 Afroasiatic, 3 Niger-Congo, 3 Turkic, 2 Austronesian, 2 Dravidian, 1 Austroasiatic, 1 Kra-Dai, 1 Tupian, 1 Sino-Tibetan), as well as 2 isolates. The full list of language \textsc{iso} 639-2 codes is reported in \cref{fig:respos}.

In order to simulate a zero-shot setting, we hold out in turn half of all possible task--language pairs 
and regard them as unseen, while treating the others as seen pairs.
The partition is performed in such a way that a held-out pair has data available for the same task in a different language, and for the same language in a different task.\footnote{We use the controlled partitioning for the following reason. If a language lacks data both for NER and for POS, the proposed factorization method cannot provide estimates for its posterior. We leave model extensions that can handle such cases for future work.} Under this constraint, pairs are assigned to train or evaluation at random.\footnote{See \cref{ssec:distsize} for further experiments on splits controlled for language distance and sample size.}

We randomly split the WikiANN datasets
into training, development, and test portions with a proportion of 80-10-10. 
We use the provided splits for UD; if the training set for a language is missing, we treat the test set as such when the language is held out, and as a training set when it is among the seen pairs.\footnote{Note that, in the second case, no evaluation takes place on such language.}

\subsection{Hyper-parameters}\label{ssec:hyper}
The multilingual \textsc{m-bert} encoder is initialized with parameters pre-trained on masked language modeling and next sentence prediction on 104 languages \cite{devlin2019bert}.\footnote{Available at \rurl{github.com/google-research/bert/blob/master/multilingual.md}} 
We opt for the cased \textsc{Bert-Base} architecture, which consists of 12 layers with 12 attention heads and a hidden size of 768. As a consequence, this is also the dimension $e$ of each encoded WordPiece unit, a subword unit obtained through BPE \cite{wu2016google}.
The dimension $h$ of the multivariate Gaussian for task and language latent variables is set to $100$. 
The deep feed-forward networks $f_\lv{\psi}$ and $f_\lv{\phi}$ 
have 6 layers with a hidden size of 400 for the first layer, 768 for the internal layers, and {ReLU} non-linear activations. Their depth and width were selected based on validation performance.

The expectations over latent variables in \cref{eq:2} are approximated through 3 Monte Carlo samples per batch during training.
The KL terms are weighted with $\frac{1}{|K|}$ uniformly across training, where $|K|$ is the number of mini-batches.\footnote{We found this weighting strategy to work better than annealing as proposed by \citet{blundell2015weight}.} 
We initialize all the means $\lv{m}$
of the variational approximation
with a random sample from $\mathcal{N}(0, 0.1)$,
and the parameters for co-variance matrices $S$
of the variational approximation
with a random sample from $\mathcal{U}(0, 0.5)$ following \citet{stoleematrix}.
We choose $k=10$ as the number of columns of $B$ so it fits into memory.
The maximum sequence length for inputs is limited to 250.
The batch size is set to 8, and the best setting for the Adam optimizer \cite{kingma2015adam} was found to be an initial learning rate of $5 \cdot 10^{-6}$ based on grid search.
In order to avoid over-fitting, we perform early stopping with a patience of 10 and a validation frequency of 2.5K steps.

\subsection{Baselines}
We consider four baselines for cross-lingual transfer that also use $\bert$ as an encoder shared across all languages. 

\paragraph{First Baseline.}
A common approach is transfer from the \textbf{nearest source} (NS) language, which selects the most compatible source to a target language in terms of similarity.
In particular, the selection can be based on family membership \cite{zeman2008cross,cotterell-heigold-2017-cross,kann-etal-2017-one}, typological features \citep{Deri-2016}, KL-divergence between part-of-speech trigram distributions \citep{Rosa:2015acl,agic2017cross}, tree edit distance of delexicalized dependency parses \citep{Ponti:2018acl}, or a combination of the above \citep{Lin:2019acl}. In our work, during evaluation, we choose the classifier associated with the observed language with the highest cosine similarity between its typological features and those of the held-out language. These features are sourced from URIEL \cite{littell2017uriel} and contain information about family, area, syntax, and phonology.

\paragraph{Second Baseline.}
We also consider transfer from the \textbf{largest source} (LS) language, i.e.\ the language with most training examples. This approach has been adopted by several recent works on cross-lingual transfer \citep[\textit{inter alia}]{conneau2018xnli,artetxe2019cross}. In our implementation, we always select the English classifier for prediction.\footnote{We include English to make the baseline more competitive, but note that this language is not available for our generative model as it is both Indo-European and resource-rich.} In order to make this baseline comparable to our model, we adjust the number of English NER training examples to the sum of the examples available for all seen languages $\mathcal{S}$.\footnote{The number of NER training examples is 1,093,184 for the first partition and 520,616 for the second partition.}

\paragraph{Third Baseline.}
Next, we apply a protocol designed by \citet{rahimi2019massively} for weighting the predictions of a classifier ensemble according to their reliability. For a specific task, the reliability of each language-specific classifier is estimated through a Bayesian graphical model.
Intuitively, this model learns from error patterns, which behave more randomly for untrustworthy models and more consistently for the others. Among the protocols proposed in the paper, we opt for \textbf{BEA} in its zero-shot, token-based version, as it achieves the highest scores in a setting comparable to the current experiment. We refer to the original paper for the details.\footnote{We implemented this model through the original code at \rurl{github.com/afshinrahimi/mmner}.}

\paragraph{Fourth Baseline.}
Finally, we take inspiration from \citet{wu2019beto}.  The \textbf{joint multilingual} (JM) baseline, contrary to the previous baselines, consists of two classifiers (one for POS tagging and another for NER) shared among all observed languages for a specific task. We follow the original implementation of \citet{wu2019beto} closely adopting all recommended hyper-parameters and strategies, such as freezing the parameters of all encoder layers below the 3\textsuperscript{rd} for sequence labeling tasks.

It must be noted that the number of parameters in our generative model scales better than baselines with language-specific classifiers, but worse than those with language-agnostic classifiers, as the number of languages grows. However, even in the second case, increasing the depth of baselines networks to match the parameter count is detrimental if the $\bert$ encoder is kept trainable, which was also verified in previous work \citep{peters2019tune}.

\section{Results and Discussion}
\label{s:results}
\begin{figure*}[p]
    \centering
    \vspace{-1cm}
    \includegraphics[width=\textwidth]{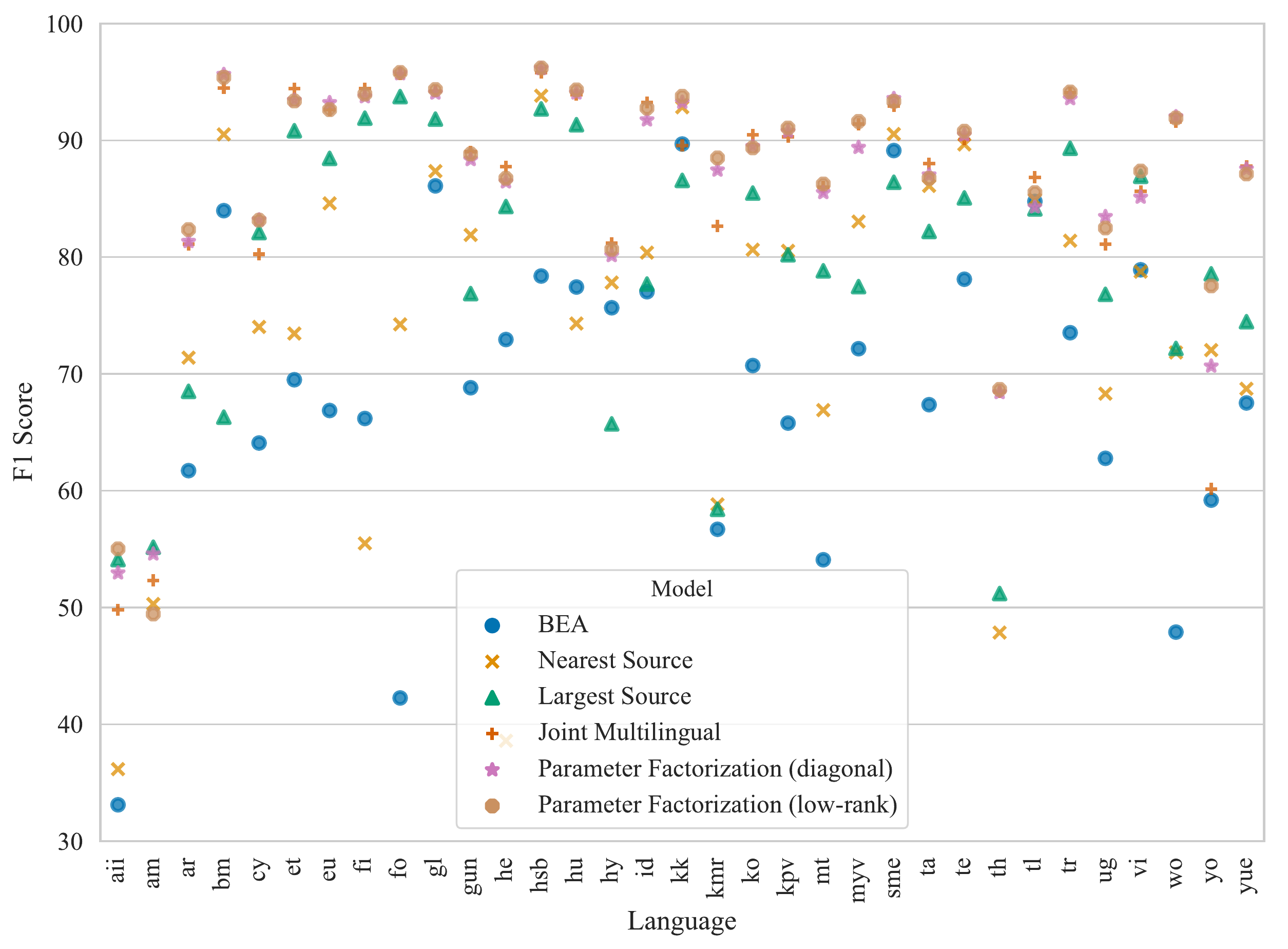}
    \includegraphics[width=\textwidth]{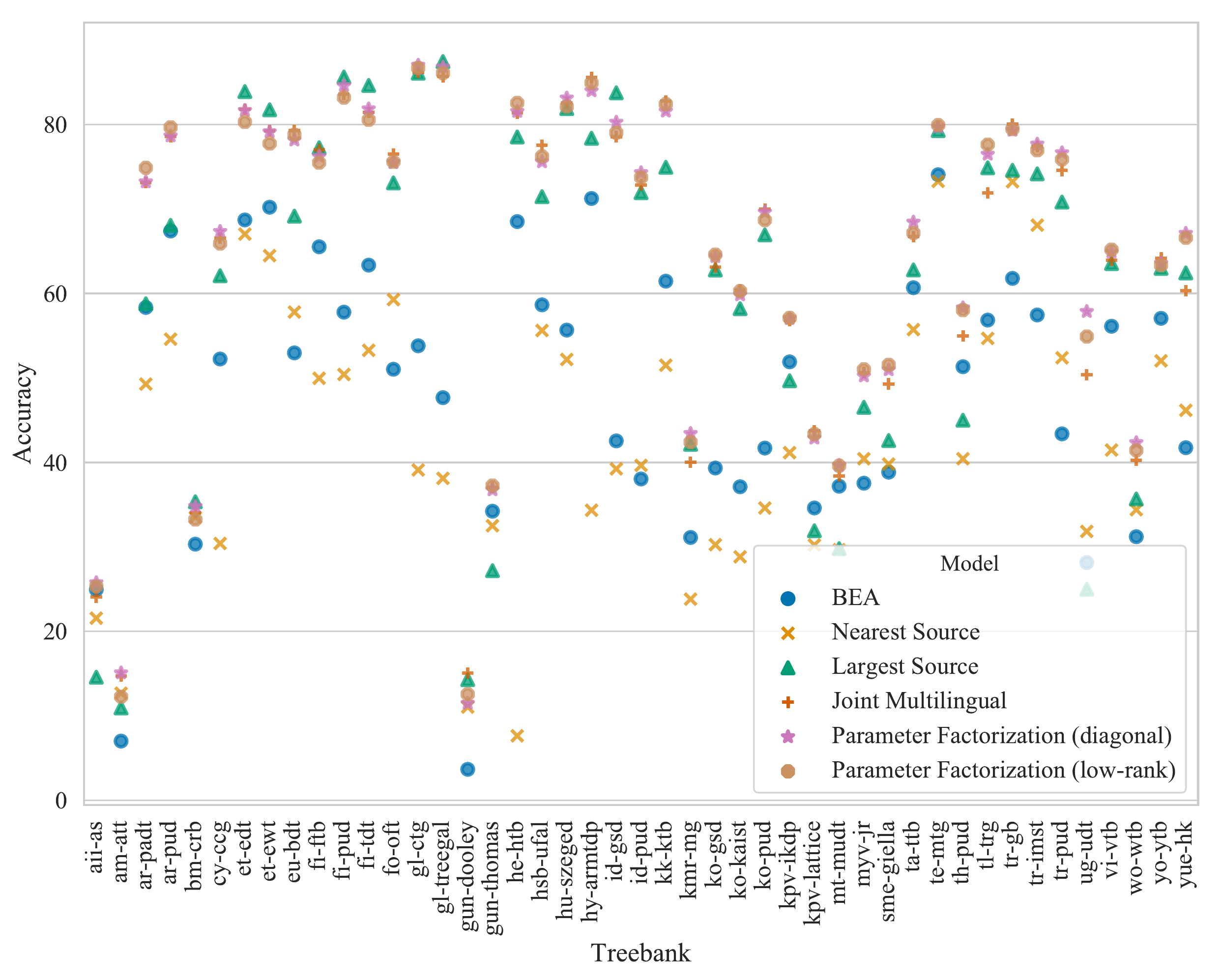}
    \vspace{-1cm}
    \caption{Results for NER (top) and POS tagging (bottom): four baselines for cross-lingual transfer compared to Matrix Factorization with diagonal co-variance and diagonal plus low-rank co-variance.}
    \label{fig:respos}
\end{figure*}

\begin{table*}[t]
    \centering
    {\small
    \begin{tabularx}{1.0\textwidth}{l YYYY|YY}
\toprule
   \textbf{Task} & \textbf{BEA} & \textbf{NS} & \textbf{LS} & \textbf{JM} & \textbf{PF-d} & \textbf{PF-lr} \\
   \hline
POS & 47.65 $\pm$ 1.54 & 42.84 $\pm$ 1.23 & 60.51 $\pm$ 0.43 & 64.04 $\pm$ 0.18 & {\bf 65.00 $\pm$ 0.12} & {\bf 64.71 $\pm$ 0.18} \\
NER & 66.45 $\pm$ 0.56 & 74.16 $\pm$ 0.56 & 78.97 $\pm$ 0.56 & 85.65 $\pm$ 0.13 & 86.26 $\pm$ 0.17 & {\bf 86.70 $\pm$ 0.10} \\
\bottomrule
    \end{tabularx}}%
    \caption{Results per task averaged across all languages.}
    \label{tab:avgres}
\end{table*}

\subsection{Zero-shot Transfer}
\label{ssec:zeroshot}
Firstly, we present the results for zero-shot prediction based on our generative model using both of the approximate inference schemes (with diagonal co-variance \textbf{PF-d} and factor co-variance \textbf{PF-lr}).
\Cref{tab:avgres} summarizes the results on the two tasks of POS tagging and NER averaged across all languages. 
Our model (in both its variants) outperforms the four baselines on both tasks, including state-of-the-art alternative methods.
In particular, PF-d and PF-lr gain 4.49 / 4.20 in accuracy (\textasciitilde 7\%) for POS tagging and 7.29 / 7.73 in F1 score (\textasciitilde 10\%) for NER on average compared to transfer from the largest source (\textbf{LS}), the strongest baseline for single-source transfer. Compared to multilingual joint transfer from multiple sources (\textbf{JM}), our two variants gain 0.95 / 0.67 in accuracy (\textasciitilde 1\%) for POS tagging and +0.61 / +1.05 in F1 score (\textasciitilde 1\%).

More details about the individual results on each task--language pair are provided in \cref{fig:respos}, which includes the mean of the results over 3 separate runs. 
Overall, we obtain improvements in 23/33 languages for NER and on 27/45 treebanks for POS tagging, which further supports the benefits of transferring both from tasks and languages. 

Considering the baselines, the relative performance of LS versus NS is an interesting finding per se. LS largely outperforms NS on both POS tagging and NER. This shows that having more data is more informative than relying primarily on similarity according to linguistic properties. This finding contradicts the received wisdom \citep[\textit{inter alia}]{Rosa:2015acl,cotterell-heigold-2017-cross,Lin:2019acl} that related languages tend to be the most reliable source. 
We conjecture that this is due to the pre-trained multi-lingual $\bert$ encoder, which helps to bridge the gap between unrelated languages \citep{wu2019beto}. 

The two baselines that hinge upon transfer from multiple sources lie on opposite sides of the spectrum in terms of performance. 
On the one hand, BEA achieves the lowest average score for NER, and surpasses only NS for POS tagging. 
We speculate that this is due to the following: i) adapting the protocol from \citet{rahimi2019massively} to our model implies assigning a separate classifier head to each task--language pair, each of which is exposed to fewer examples compared to a shared one. 
This fragmentation fails to take advantage of the massively multilingual nature of the encoder; ii) our language sample is more typologically diverse, which means that most source languages are unreliable predictors. 
On the other hand, JM yields extremely competitive scores. 
Similarly to our model, it integrates knowledge from multiple languages and tasks. The extra boost in our model stems from its ability to disentangle each aspect of such knowledge and recombine it appropriately.

\begin{table}[!t]
    \centering
    {\small
    \begin{tabularx}{1.0\columnwidth}{l YY|YY}
\toprule
   \textbf{Task} & \multicolumn{2}{c|}{$|L|=11$} & \multicolumn{2}{c}{$|L|=22$} \\
   & Sim & Dif & Sim & Dif \\
   \hline
   POS & 72.44 & 53.25 & 66.59 & 63.22\\
   NER & 89.51 & 81.73 & 86.78 & 85.12\\

\bottomrule
    \end{tabularx}}%
    \caption{Average performance when relying on $|L|$ similar (\textit{Sim}) versus different (\textit{Dif}) languages in the train and evaluation sets.}
    \label{tab:affinity}
\end{table}

Moreover, comparing the two approximate inference schemes from \cref{ssec:svi}, PF-lr obtains a small but statistically significant improvement over PF-d in NER, whereas they achieve the same performance on POS tagging. This means that the posterior is modeled well enough by a Gaussian where co-variance among co-variates is negligible.

We see that even for the best model (PF-lr) there is a wide variation in the scores for the same task across languages. POS tagging accuracy ranges from $12.56 \pm 4.07$ in Guaran\'i to $86.71 \pm 0.67$ in Galician, and NER F1 scores range from $49.44 \pm 0.69$ in Amharic to $96.20 \pm 0.11$ in Upper Sorbian. Part of this variation is explained by the fact that the multilingual $\bert$ encoder is not pre-trained in a subset of these languages (e.g., Amharic, Guaran\'i, Uyghur). Another cause is more straightforward: the scores are expected to be lower in languages for which we have fewer training examples in the seen task--language pairs.

\subsection{Language Distance and Sample Size}
\label{ssec:distsize}
While we designed the language sample to be both realistic and representative of the cross-lingual variation, there are several factors inherent to a sample that can affect the zero-shot transfer performance: i) \textit{language distance}, the similarity between seen and held-out languages; and ii) \textit{sample size}, the number of seen languages. In order to disentangle these factors, we construct subsets of size $|L|$ so that training and evaluation languages are either maximally similar (\textit{Sim}) or maximally different (\textit{Dif}). As a proxy measure, we consider as `similar' languages belonging to the same family. In \cref{tab:affinity}, we report the performance of parameter factorization with diagonal plus low-rank co-variance (PF-lr), the best model from \cref{ssec:zeroshot}, for each of these subsets.

Based on \cref{tab:affinity}, there emerges a trade-off between language distance and sample size. In particular, performance is higher in \textit{Sim} subsets compared to \textit{Dif} subsets for both tasks (POS and NER) and for both sample sizes $|L| \in \{11, 22\}$. In larger sample sizes, the average performance increases for \textit{Dif} but decreases for \textit{Sim}. Intuitively, languages with labeled data for several relatives benefit from small, homogeneous subsets. Introducing further languages introduces noise. Instead, languages where this is not possible (such as isolates) benefit from an increase in sample size.

\begin{figure*}[t]
    \centering
    \includegraphics[width=0.98\textwidth, trim={0 13mm 0 0}]{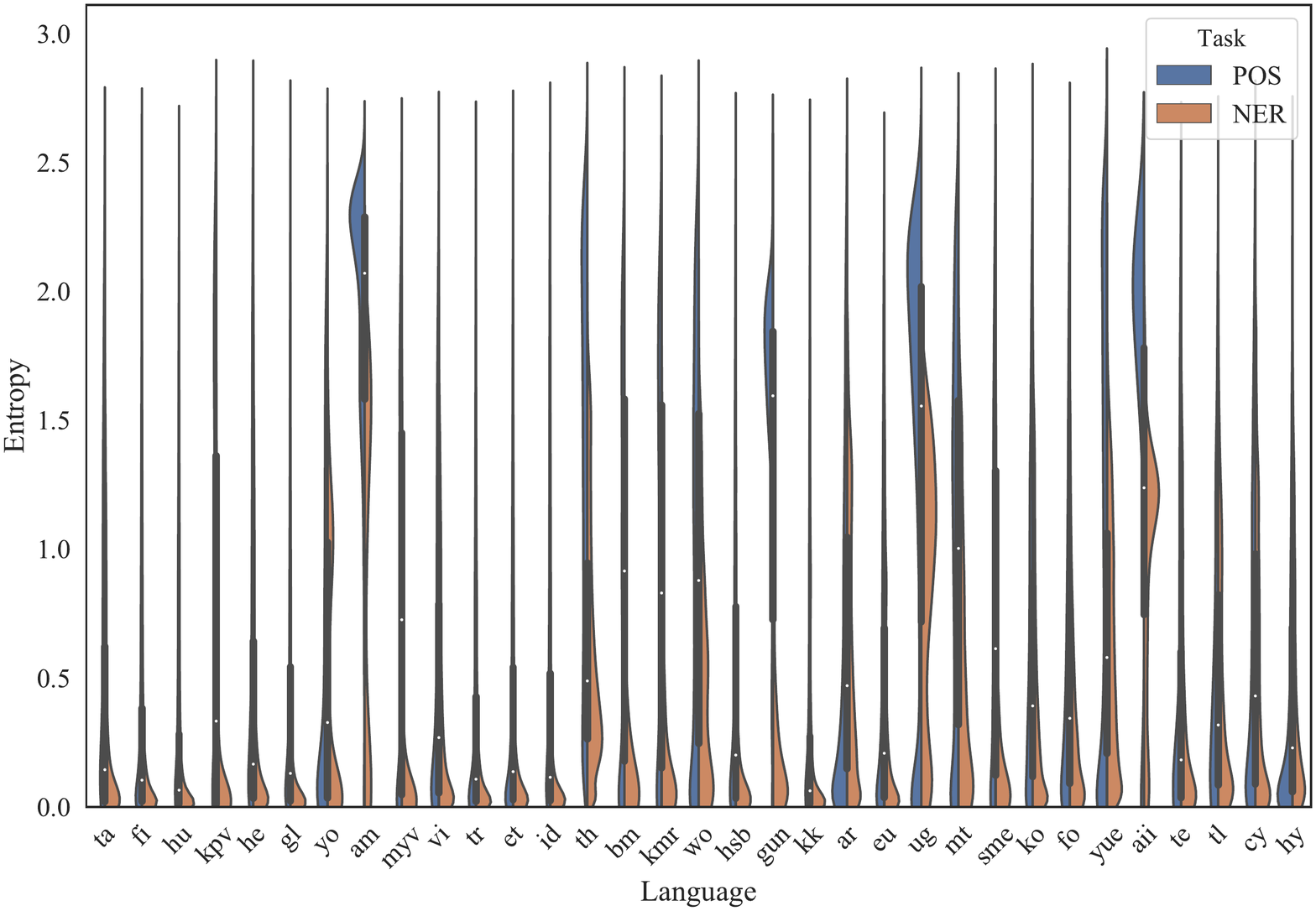}
    \vspace{6.pt}
    \caption{Entropy of the posterior predictive distributions over classes for each test example. The higher the entropy, the more uncertain the prediction.}
    \label{fig:predentropy}
\end{figure*}

\subsection{Entropy of the Predictions}
A notable problem of point estimate methods is their tendency to assign most of the probability mass to a single class even in scenarios with high uncertainty. Zero-shot transfer is one of such scenarios, because it involves drastic distribution shifts in the data \cite{rabanser2019failing}. A key advantage of Bayesian inference, instead, is marginalization over parameters, which yields smoother posterior predictive distributions \cite{Kendall:2017nips,wilson2019bayesian}.

We run an analysis of predictions based on (approximate) Bayesian model averaging. First, we randomly sample 800 examples from each test set of a task--language pair. For each example, we predict a distribution over classes $Y$ through model averaging based on 10 samples from the posteriors. We then measure the prediction entropy of each example, i.e.\ $\mathrm{H}(p) = -\sum_y^{|Y|} p(Y=y) \ln p(Y=y)$, whose plot is shown in \cref{fig:predentropy}.

Entropy is a measure of uncertainty.
Intuitively, the uniform categorical distribution (maximum uncertainty) has the highest entropy, 
whereas if the whole probability mass falls into a single class (maximum confidence), then the entropy $\mathrm{H} = 0$.\footnote{The maximum entropy is $\approx 2.2$ for 9 classes as in NER and $\approx 2.83$ for 17 classes as in POS tagging.}
As it emerges from \cref{fig:predentropy}, predictions in certain languages tend to have higher entropy on average, such as in Amharic, Guaran{\'i}, Uyghur, or Assyrian Neo-Aramaic. This aligns well with the performance metrics in \cref{fig:respos}. In practice, languages with low scores tend to display high entropy in the predictive distribution, as expected. 
To verify this claim, we measure the Pearson's correlation between entropies of each task--language pair in \cref{fig:predentropy} and performance metrics. We find a very strong negative correlation with a coefficient of $\rho = -0.914$ and a two-tailed p-value of $1.018 \times 10^{-26}$.

\section{Related Work}
\label{s:related}
Our approach builds on ideas from several different fields: cross-lingual transfer in NLP, with a particular focus on sequence labeling tasks, as well as matrix factorization, contextual parameter generation, and neural Bayesian methods.

\paragraph{Cross-Lingual Transfer for Sequence Labeling.} 
One of the two dominant approaches for cross-lingual transfer is \textit{projecting annotations} from a source language text to a target language text. This technique was pioneered by \newcite{yarowsky2001inducing} and \newcite{hwa-2005} for parsing, and later extended to applications such as POS tagging \cite{Das:2011acl,Garrette:2013acl,Tackstrom:2012naacl,Duong:2014emnlp,Huck:2019ws} 
and NER \cite{Ni:2017acl,Enghoff:2018ws,Agerri:2018lrec,Jain:2019emnlp}. This requires tokens to be aligned through a parallel corpus, a machine translation system, or a bilingual dictionary \cite{Durrett:2012emnlp,Mayhew:2017emnlp}. However, creating machine translation and word-alignment systems demands parallel texts in the first place, while automatically induced bilingual lexicons are noisy and offer only limited coverage  \cite{Artetxe:2018iclr,Duan:2020acl}. Furthermore, errors inherent to such systems cascade along the projection pipeline \citep{Agic:2015acl}. 

The second approach,\textit{ model transfer}, offers higher flexibility \cite{conneau2018xnli}. The main idea is to train a model directly on the source data, and then deploy it onto target data \citep{zeman2008cross}. Crucially, bridging between different lexica requires input features to be language-agnostic. While originally this implied delexicalization, replacing words with universal POS tags \cite{McDonald:2011emnlp,Dehouck:2017eacl}, cross-lingual Brown clusters \cite{Tackstrom:2012naacl,Rasooli:2017tacl}, or cross-lingual knowledge base grounding through wikification \cite{Camacho:2016nasari,Tsai:2016conll}, more recently these have been supplanted by cross-lingual word embeddings \cite{Ammar:2016tacl,zhang-etal-2016-ten,Xie:2018emnlp,Ruder:2019jair} and multilingual pretrained language models \cite{devlin2019bert,Conneau:2020acl}. 

An orthogonal research thread regards the \textit{selection of the source language(s)}.
In particular, multi-source transfer was shown to surpass single-best source transfer in NER \cite{Fang:2017acl,rahimi2019massively} and POS tagging \cite{Enghoff:2018ws,Plank:2018emnlp}. Our parameter space factorization model can be conceived as an extension of multi-source cross-lingual model transfer to a cross-task setting.

\paragraph{Data Matrix Factorization.}
Although we are the first to propose a factorization of the \textit{parameter} space for unseen combinations of tasks and languages, the factorization of \textit{data} for collaborative filtering and social recommendation is an established research area. In particular, the missing values in sparse data structures such as user-movie review matrices can be filled via probabilistic matrix factorization (PMF) through a linear combination of user and movie matrices \cite[\textit{inter alia}]{mnih2008probabilistic,ma2008sorec,shan2010generalized} or through neural networks \citep{dziugaite2015neural}. Inference for PMF can be carried out through MAP inference \citep{dziugaite2015neural}, Markov chain Monte Carlo \cite[MCMC;][]{salakhutdinov2008bayesian} or stochastic variational inference \cite{stoleematrix}. Contrary to prior work, we perform factorization on latent variables (task- and language-specific parameters) rather than observed ones (data).

\paragraph{Contextual Parameter Generation.} Our model is reminiscent of the idea that parameters can be conditioned on language representations, as proposed by \citet{Platanios:2018emnlp}. However, since this approach is limited to a single task and a joint learning setting, it is not suitable for generalization in a zero-shot transfer setting.

\paragraph{Bayesian Neural Models.} So far, these models have found only limited application in NLP for resource-poor languages, despite their desirable properties. Firstly, they can incorporate priors over parameters to endow neural networks with the correct inductive biases towards language: \citet{Ponti:2019emnlp} constructed a prior imbued with universal linguistic knowledge for zero- and few-shot character-level language modeling. Secondly, they avoid the risk of over-fitting by taking into account uncertainty. For instance, \citet{Shareghi:2019naacl} and
\citet{doitch2019perturbation} use a perturbation model to sample high-quality and diverse solutions for structured prediction in cross-lingual parsing.

\section{Conclusion}
\label{s:conclusions}
The main contribution of our work is a Bayesian generative model for multiple NLP tasks and languages. 
At its core lies the idea that the space of neural weights can be factorized into latent variables for each task and each language. While training data are available only for a meager subset of task--language combinations, our model opens up the possibility to perform prediction in novel, undocumented combinations at evaluation time.
We performed inference through stochastic variational methods, and ran experiments on zero-shot named entity recognition (NER) and part-of-speech (POS) tagging in a typologically diverse set of 33 languages. 
Based on the reported results, we conclude that leveraging the information from tasks and languages simultaneously is superior to model transfer from English (relying on more abundant in-task data in the source language), from the most typologically similar language (relying on prior information on language relatedness), or from multiple source languages. Moreover, we found that the entropy of predictive posterior distributions obtained through Bayesian model averaging correlates almost perfectly with the error rate in the prediction.
As a consequence, our approach holds promise to alleviating data paucity issues for a wide spectrum of languages and tasks, and to make knowledge transfer more robust to uncertainty.

Finally, we remark that our model is amenable to be extended to multilingual tasks beyond sequence labeling---such as natural language inference \citep{conneau2018xnli} and question answering \citep{artetxe2019cross,Lewis:2019arxiv,tydiqa}---and to zero-shot transfer across combinations of multiple modalities (e.g.\ speech, text, and vision) with tasks and languages. We leave these exciting research threads for future research. 

\section*{Acknowledgements}
We would like to thank action editor Jacob Eisenstein and the three anonymous reviewers at TACL. 
This work is supported by the ERC Consolidator Grant LEXICAL (no 648909) and the Google Faculty Research Award 2018. RR was partially funded by ISF personal grants No. 1625/18.

\bibliography{references}
\bibliographystyle{acl_natbib}

\clearpage
\appendix
\section{KL-divergence of Gaussians}
\label{app:kldivgauss}

If both $p \triangleq \mathcal{N}(\lv{\mu}, {\Sigma})$ and $q \triangleq \mathcal{N}(\lv{m}, {S})$ are multivariate Gaussians, their KL-divergence can be computed analytically as follows:

\begin{align}
\begin{split}
\KLD{q}{p} &= \frac{1}{2} \Bigl[ \ln \frac{|{S}|}{|{\Sigma}|} -d + \Tr({S}^{-1} {\Sigma} ) \\
&+ (\lv{m} - \lv{\mu})^{\top} {S}^{-1} (\lv{m} - \lv{\mu}) \Bigr]
\end{split}
\end{align}

\noindent
By substituting $\lv{m} = \lv{0}$ and ${S} = {I}$, it is trivial to obtain \cref{eq:klnormal}.

\section{Visualization of the Learned Posteriors}
\begin{figure}[h]
    \centering
    \includegraphics[width=\columnwidth, trim={0 15mm 0 0}]{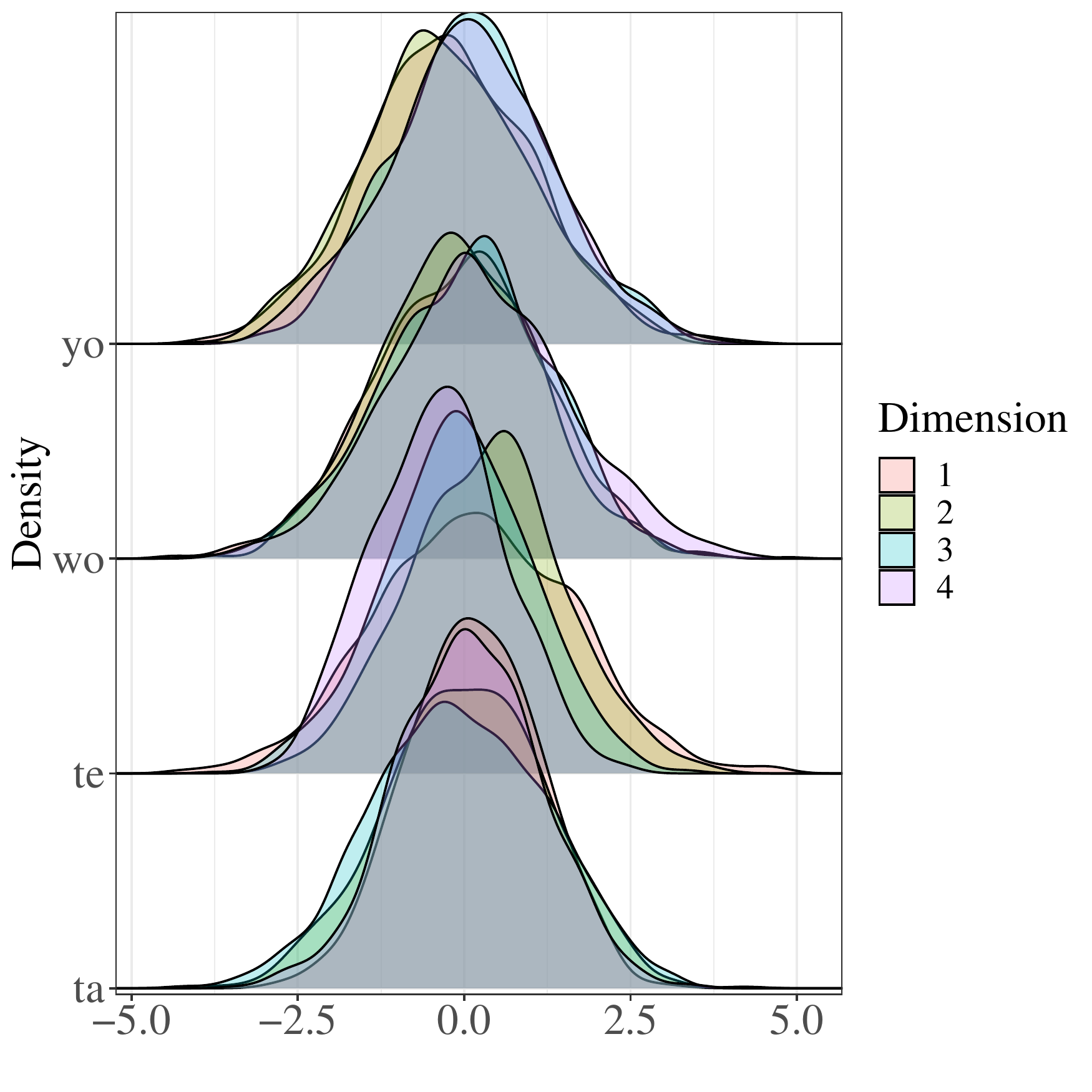}
    \caption{Samples from the posteriors of 4 languages, PCA-reduced to 4 dimensions.}
    \label{fig:langpost}
\end{figure}

\noindent
The approximate posteriors of the latent variables can be visualized in order to study the learned representations for languages.
Previous work \cite{johnson2017google,ostling2017continuous,malaviya2017learning,bjerva2018phonology} induced point estimates of language representations from artificial tokens concatenated to every input sentence,
or from the aggregated values of the hidden state of a neural encoder. The information contained in such representations depends on the task \citep{bjerva2018phonology}, but mainly reflects the structural properties of each language \cite{Bjerva:2019cl}.

In our work, due to the estimation procedure, languages are represented by full distributions rather than point estimates.
By inspecting the learned representations, language similarities do not appear to follow the structural properties of languages.
This is most likely due to the fact that parameter factorization takes place \textit{after} the multi-lingual $\bert$ encoding, which blends the structural differences across languages. A fair comparison with previous works without such an encoder is left for future investigation.

As an example, consider two pairs of languages from two distinct families: Yoruba and Wolof are Niger-Congo from the Atlantic-Congo branch,
Tamil and Telugu are Dravidian. We take 1,000 samples from the approximate posterior over the latent variables for each of these languages. In particular, we focus on the variational scheme with a low-rank co-variance structure.
We then reduce the dimensionality of each sample to 4 through PCA,\footnote{Note that the dimensionality reduced samples are also Gaussian since PCA is a linear method.}
and we plot the density along each resulting dimension in \cref{fig:langpost}. 
We observe that density areas of each dimension do not necessarily overlap between members of the same family. Hence, the learned representations depend on more than genealogy.

\end{document}